\documentclass{article}

\usepackage[preprint,nonatbib]{confstyle}

\usepackage[utf8]{inputenc} 
\usepackage[T1]{fontenc}    
\usepackage{hyperref}       
\usepackage{url}            
\usepackage{booktabs}       
\usepackage{amsfonts}       
\usepackage{nicefrac}       
\usepackage{microtype}      
\usepackage{xcolor}         
\usepackage{graphicx}  
\usepackage{setspace}
\usepackage{multirow}
\usepackage{hyperref}

\usepackage{mathptmx} 
\usepackage{times} 
\usepackage{amssymb}  

\usepackage{listings} 

\usepackage{algorithm}  
\usepackage{algorithmic}
\usepackage{amsmath}

\usepackage{balance}
\usepackage{adjustbox}
\usepackage{tabularx}

\usepackage{subcaption} 
\usepackage{tcolorbox}
\usepackage{wrapfig}

\tcbuselibrary{breakable}
\tcbuselibrary{listings}

\title{Evaluating Multimodal Large Language Models with Daily Composite Tasks in Home Environments}

\author{Zhenliang Zhang$^{1*}$, Yuxi Wang$^{1,2*}$, Hongzhao Xie$^{1}$, Shiyun Zhao$^{1}$, Mingyuan Liu$^{1}$, \\ \textbf{Yujie Lu$^{1}$, Xinyi He$^{1,3}$, Zhenku Cheng$^{2}$, Yujia Peng$^{2,3,1}$}
\thanks{$^{*}$ equal contributions. {\raggedright Correspondence to: Yujia Peng \href{mailto:yujia_peng@pku.edu.cn}{yujia\_peng@pku.edu.cn}}. Affliations: $^{1}$ State Key Laboratory of General Artificial Intelligence, Beijing Institute for General Artificial Intelligence, Beijing, China. $^{2}$ School of Psychological and Cognitive Sciences and Beijing Key Laboratory of Behavior and Mental Health, Key Laboratory of Machine Perception (Ministry of Education), Peking University, Beijing, China. $^{3}$ School of Intelligence Science and Technology, Peking University, Beijing 100871, China.\\
}
}

\makeatletter
\def\thanks#1{\protected@xdef\@thanks{\@thanks
        \protect\footnotetext{#1}}}
\makeatother

\begin{document}

\maketitle

\begin{abstract}
A key feature differentiating artificial general intelligence (AGI) from traditional AI is that AGI can perform composite tasks that require a wide range of capabilities. Although embodied agents powered by multimodal large language models (MLLMs) offer rich perceptual and interactive capabilities, it remains largely unexplored whether they can solve composite tasks. In the current work, we designed a set of composite tasks inspired by common daily activities observed in early childhood development. Within a dynamic and simulated home environment, these tasks span three core domains: object understanding, spatial intelligence, and social activity. We evaluated 17 leading proprietary and open-source MLLMs on these tasks. The results consistently showed poor performance across all three domains, indicating a substantial gap between current capabilities and general intelligence requirements. Together, our tasks offer a preliminary framework for evaluating the general capabilities of embodied agents, marking an early but significant step toward the development of embodied MLLMs and their real-world deployment. 
\end{abstract}

\section{Introduction}

With the release of the generative pre-trained transformer (GPT) model series \cite{brown2020language,achiam2023gpt,radford2019language}, artificial general intelligence (AGI) has reemerged as a focal point in AI research. Recent foundational models have demonstrated near-human-level capabilities in domains such as natural language processing \cite{achiam2023gpt}, image segmentation \cite{kirillov2023segment}, and robotics \cite{driess2023palm}. Nevertheless, it remains unclear whether existing foundation models can be considered as forms of AGI. A central point of debate is how to define and evaluate whether an AI system qualifies as Artificial General Intelligence. On a theoretical level, multiple research teams, including OpenAI and Google DeepMind, have attempted to define AGI rating systems \cite{morris2024position}. These efforts provide important references, they have yet to establish a concrete, complete definition, or quantifiable evaluation standards. As a result, translating theoretical principles into practical AGI evaluations remains an unresolved challenge. Here, we argue that a key distinction of AGI, compared to traditional AI, lies in its ability to perform composite tasks that require a broad spectrum of capabilities \cite{peng2024tong}, namely \textbf{general abilities}. It is therefore essential to develop evaluations that define and quantify both the extent and the specific dimensions in which current systems fall short of achieving AGI. 

Classic AI evaluation approaches have limitations when applied to AGI evaluation. Taking the Turing Test \cite{oppy2003turing} as an example, the underlying core premise is determining whether a machine exhibits human-level intelligence. However, the test heavily relies on the knowledge and cognitive abilities of human judges, bearing limited objectivity and standardization. Moreover, the Turing Test has been repeatedly beaten by chatbots (e.g., Google Duplex Voice AI), which inherit carefully designed response algorithms rather than genuine intelligence. Another common approach is task-oriented benchmark testing, where AI algorithms are evaluated on specific datasets for predefined tasks. For example, numerous datasets (e.g., ImageNet \cite{deng2009imagenet}, COCO\cite{lin2014microsoft}, and Visual Question Answering (VQA) \cite{antol2015vqa}) have emerged in the past decades, providing foundational resources for research and development. However, classic benchmarks share limitations of over-specialization and propensity to overfitting. First, most classic benchmarks focus solely on solving specific problems, making them unsuitable for assessing AGI. Second, with the release of a fixed benchmark, models were optimized for specific datasets, resulting in overfitting, where systems perform well in controlled settings but fail in complex and open-ended real-world scenarios.

\begin{wrapfigure}{r}{0.5\textwidth}
    \centering
    \includegraphics[width=0.49\textwidth]{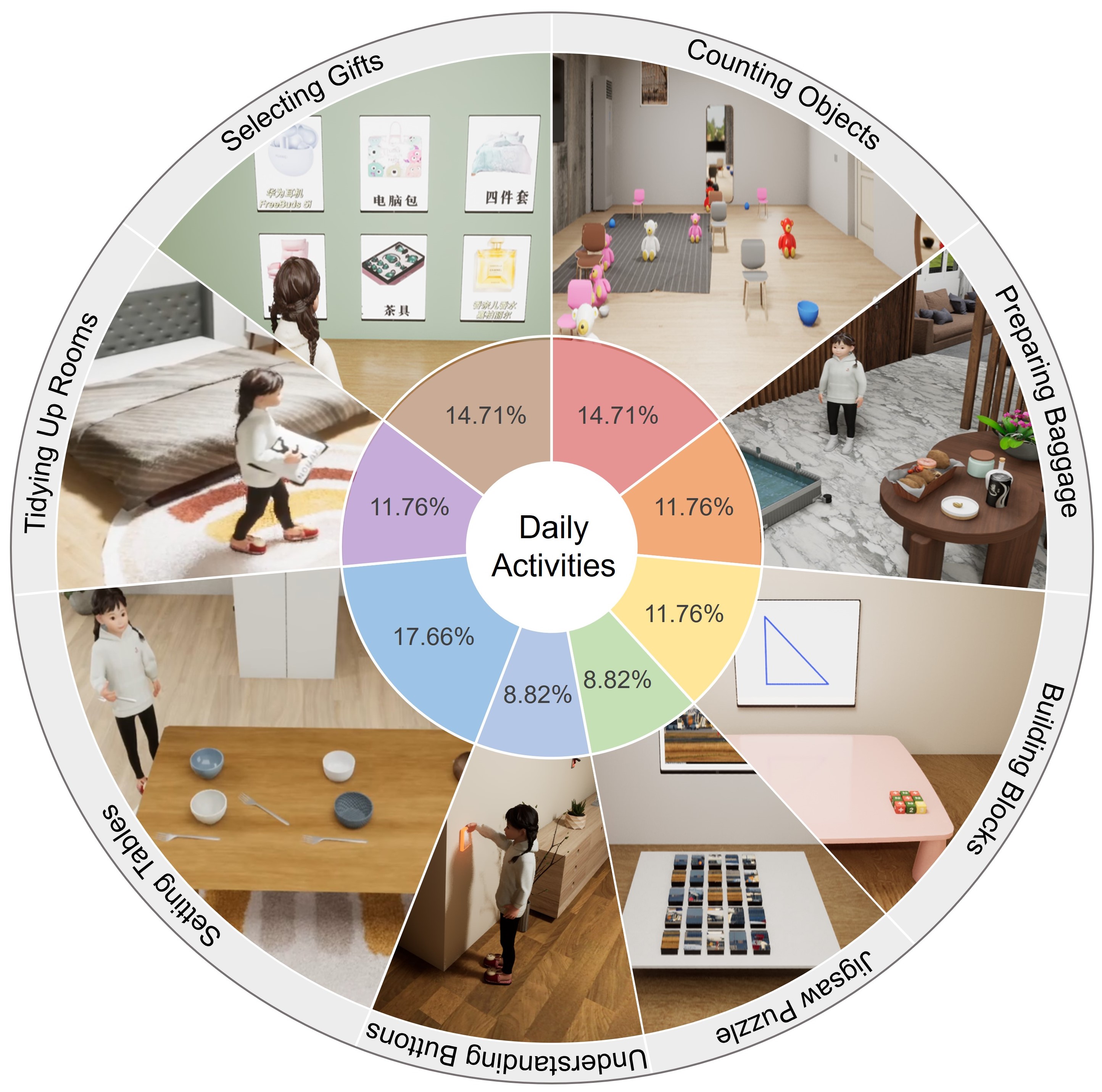}
    \caption{The illustration of eight task types.}
    \label{fig:tasks}
\end{wrapfigure}

In recent years, multi-task evaluation platforms such as MMBench \cite{liu2024mmbench} and FlagEval \cite{he2025flagevalmm} have emerged, offering cross-modal and multi-dimensional testing that is both scalable and easy to deploy. These frameworks cover a broad spectrum of cognitive and perceptual tasks, providing valuable insights into model performance across various domains. However, a significant limitation of such platforms is their lack of embodied evaluation capabilities—such as interactive virtual environments or real-world robotic interactions—which are essential for assessing situated reasoning and physical task execution. Consequently, despite their breadth, there remains no comprehensive AGI assessment framework that integrates embodied interaction with cross-domain reasoning to effectively guide future AI development toward general intelligence. 

In the present work, we argue that the evaluation of AGI requires an embodied environment, in which an agent must continuously adapt to a dynamic world involving both physical and social interactions, while managing an unbounded set of composite tasks spanning a broad spectrum of capabilities \cite{peng2024tong}.  To structure these composite tasks, we draw inspiration from the multidimensional capabilities observed in human development from birth to age six. Foundational capabilities include visual functions such as object recognition, visual navigation for localization and movement guidance, motor skills like walking and grasping that support environmental exploration, and language processing, which enables interaction with caregivers and facilitates social feedback. These core competencies provide the groundwork for an agent to engage with the world and integrate into human society, supporting ongoing learning and development processes. 

Drawing on developmental psychology frameworks in human early childhood \cite{scharf2016developmental,taanila2005infant,spelke2007core}, we designed a set of composite tasks to evaluate 17 leading proprietary and open-source MLLM-based agents. The tasks were based on eight common daily activities in an everyday home setting,  covering three core domains: object understanding, spatial intelligence, and social activity. The task set incorporates three key design principles: (1) \textbf{Alignment with child development}: leveraging milestones from early childhood as a robust reference for intellectual and interactive complexity. (2) \textbf{Embodied composite tasks}: each task required composite abilities that combine low-level perceptual or motor abilities with high-level cognitive reasoning within an embodied environment. (3) \textbf{Ability-oriented evaluation}: the tasks were structured to systematically assess how proficient the agent is in performing a range of cognitive and interactive tasks necessary for real-world scenarios with dynamic physical and social demands.

The contributions of the present work were threefold. First, we developed a set of embodied tasks grounded in daily activities from early childhood, designed to assess general cognitive and interactive abilities within a simulated home environment. Second, we proposed an integrated evaluation pipeline that encompassed perception, reasoning, and action, enabling systematic and comprehensive analysis of MLLM-based agents in complex, interactive settings. Last but not least, we conducted extensive evaluations of the general abilities of state-of-the-art MLLMs, providing insights into the associated  limitations and implications for future development toward AGI.

\section{Related Works}

\paragraph{MLLMs}
State-of-the-art MLLMs are capable of integrating multimodal information, particularly visual input through visual language models (VLMs) \cite{szot2025multimodal}, enabling them to interact with the external world in both physical and social contexts \cite{sarch2024vlm,zhang2024navid}. However, most existing models are primarily trained for specific tasks such as manipulation\cite{gao2024multi, yang2025instructvla}, navigation \cite{zhang2024navid,nie2025wmnav}, or performance in complex benchmark suites \cite{yang2025embodiedbench}, with limited capabilities to tackle composite tasks that combine perception, reasoning, motor control, and social interaction, mirroring real-world challenges where diverse cognitive and physical skills must be orchestrated in contextually rich and dynamic environments. Thus, the task-specific training of MLLMs may limit the agents' adaptability and generalization to open-ended environments, which is essential for broader real-world applications.

Recent efforts have sought to address this challenge through the distillation of general knowledge into embodied agents \cite{sarch2024vlm}. However, the evaluation of general abilities to solve composite tasks remains underdeveloped, hindering a clear assessment of existing MLLM agents' capabilities in the real world. Consequently, a systematic framework employing composite tasks that reflect realistic daily environments is essential for the rigorous evaluation of their progress in gauging how far these agents remain from acquiring the general-purpose capabilities required for AGI.

\paragraph{MLLM evaluations}
Existing benchmarks for MLLM agents have primarily focused on tasks at either the low or high end of the complexity spectrum. Low-level tasks include manipulation \cite{zheng2022vlmbench} and navigation \cite{khanna2024goatbenchbenchmarkmultimodallifelong}, while high-level tasks range from household activities \cite{shridhar2020alfred,li2023behavior,szot2023large} to complex cognitive tasks such as spatial reasoning \cite{li202511plus, yin2025spatial} and decision making \cite{li2024embodied}. Additionally, some benchmarks assess agent performance across a range of task complexities \cite{cheng2025embodiedeval,yang2025embodiedbench}. Although these benchmarks provided valuable insight into the capabilities of MLLM agents to perform specific, well-defined tasks in controlled environments, many of these evaluations were constrained by their specific scope, often overlooking broader general abilities required for real-world competence. 

However, few evaluations have been developed to assess the general abilities of embodied agents. The ARC-AGI benchmark series \cite{chollet2024abstraction, chollet2019measure} focused on measuring general intelligence through fluid intelligence, defined as the efficiency of skill acquisition on previously unknown tasks. However, these benchmarks were designed to assess general intelligence in a broader context, using highly abstract and two-dimensional games, and do not focus on capturing the complexities and dynamics of embodied interactions in realistic environments. In contrast, drawing inspiration from child development, our task set centered on home environments as a structured platform for evaluating the general abilities of MLLM agents. This approach offered a more ecologically valid framework for assessing their capabilities in real-world scenarios. 

Despite these existing benchmarks offer important insight into agent performance on individual tasks, there is a growing need to evaluate the general abilities of MLLM agents. Drawing inspiration from child development, our task set centers on home environments as a testbed for evaluating the general abilities of MLLM agents, offering a more ecologically valid framework for assessing their capabilities in real-world scenarios. 

\paragraph{Embodied AI Platforms} 
To evaluate the performance of MLLMs agents on embodied interaction tasks, it is essential to employ embodied artificial intelligence simulation platforms. Platforms such as AI2-THOR~\cite{kolve2017ai2} and Habitat~\cite{savva2019habitat,szot2021habitat,puig2023habitat} focus on building simulated environments customized for robotic interaction tasks, while others such as ThreeDWorld~\cite{gan2020threedworld} and OmniGibson~\cite{li2023behavior} emphasize high-fidelity physical simulation to better assess agent capabilities. Furthermore, household simulation environments such as VirtualHome~\cite{puig2018virtualhome}, as well as large-scale urban simulations such as GRUTopia~\cite{wang2024grutopia} and Virtual Community~\cite{zhou2025virtual}, extend the evaluation of embodied intelligence into multi-agent settings. At the same time, virtual reality (VR) interaction has also emerged as a significant evaluation method within the AI field, exemplified by tools such as VRGym~\cite{xie2019vrgym} and VRKitchen~\cite{gao2019vrkitchen}. As a result, the assessment of intelligent agents has entered a phase characterized by symmetrical reality~\cite{zhang2024emergence, zhang2019symmetrical}, laying the foundation for comprehensive evaluations of agents designed to integrate into human social life.

\section{Methods}
\subsection{Task Design}
We introduced eight types of embodied composite tasks to evaluate the performance of MLLMs (Fig.~\ref{fig:tasks}), including: counting objects, building blocks, jigsaw puzzle, understanding buttons, setting tables, tidying up rooms, preparing baggage, and selecting gifts. These tasks were grounded in real experiences from early childhood and reflect key cognitive and motor abilities that typically emerge during this developmental stage \cite{blakey2019causality,scharf2016developmental,smidts2018object,sheldrick2019establishing,levine2012early,mulyana2022effect}.  

To organize the evaluation, we categorized the tasks into three core domains: the \textbf{object understanding} domain, including counting objects and selecting gifts; the \textbf{spatial intelligence} domain, including building blocks, jigsaw puzzles, and understanding buttons; and the \textbf{social activity} domain, including setting tables, tidying up rooms, and packing luggage. The three domains captured distinct but complementary facets of general abilities in embodied agents. The overview of each task is shown  in Fig. \ref{fig:tasks_structure}. See Appendix~\ref{AP:A} for more details.

\begin{figure*}[tb]
    \centering
    \includegraphics[width=1\linewidth]{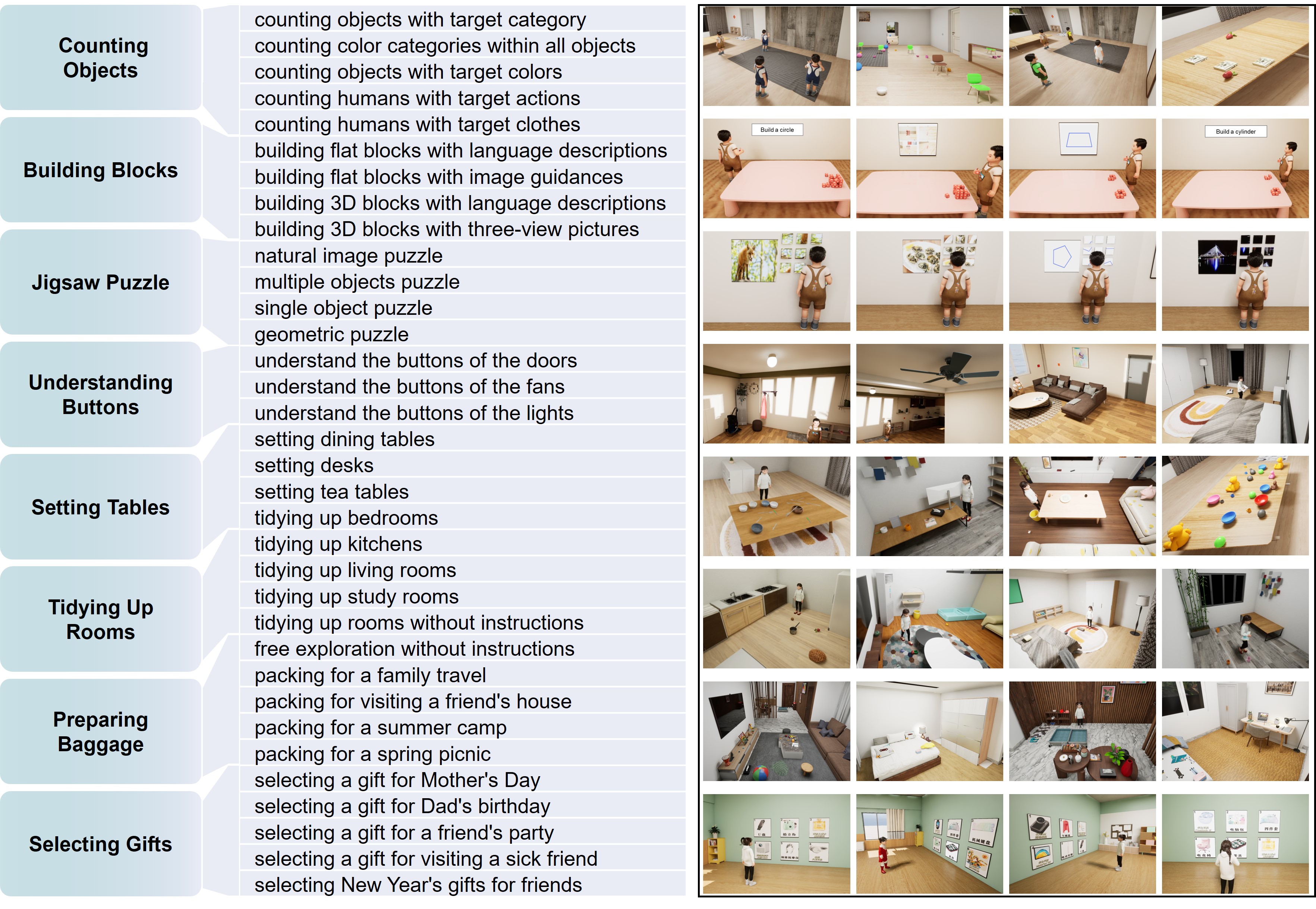}
    \caption{Overview of the eight task types with representative subtasks.}
    \label{fig:tasks_structure}
\end{figure*}

\subsection{MLLM-based Embodied Agents }

In the MLLM-driven embodied agent framework, we established a perception–reasoning–action loop in a high-fidelity 3D simulator. The MLLM agents interacted with the environment by calling pre-defined function interfaces.

\textbf{Problem Definition.} At each timestep $t$, the MLLM agent receives an observation: $o_t = \{I_t, \Sigma_t\}$, where $I_t = [I_t^{-45^\circ} \,\|\, I_t^{0^\circ} \,\|\, I_t^{+45^\circ}]$ denotes the concatenation of three RGB views (combining the left-front $45^\circ$, frontal, right-front $45^\circ$ views for a broader vision) and $\Sigma_t$ is a JSON-encoded scene description, with each object indexed by a unique \texttt{object\_id} (e.g., \texttt{object\_desk\_01}) and annotated with \texttt{name}, \texttt{color}, \texttt{position}, and \texttt{type}. Given a natural-language goal $g$, the policy $\pi_\theta$ from the MLLM outputs a symbolic action $a_t$ (API name and arguments), which is executed in the environment to produce $o_{t+1}$ until a termination condition is met.
 
\begin{figure}[tb]
    \centering
    \includegraphics[width=0.7\linewidth]{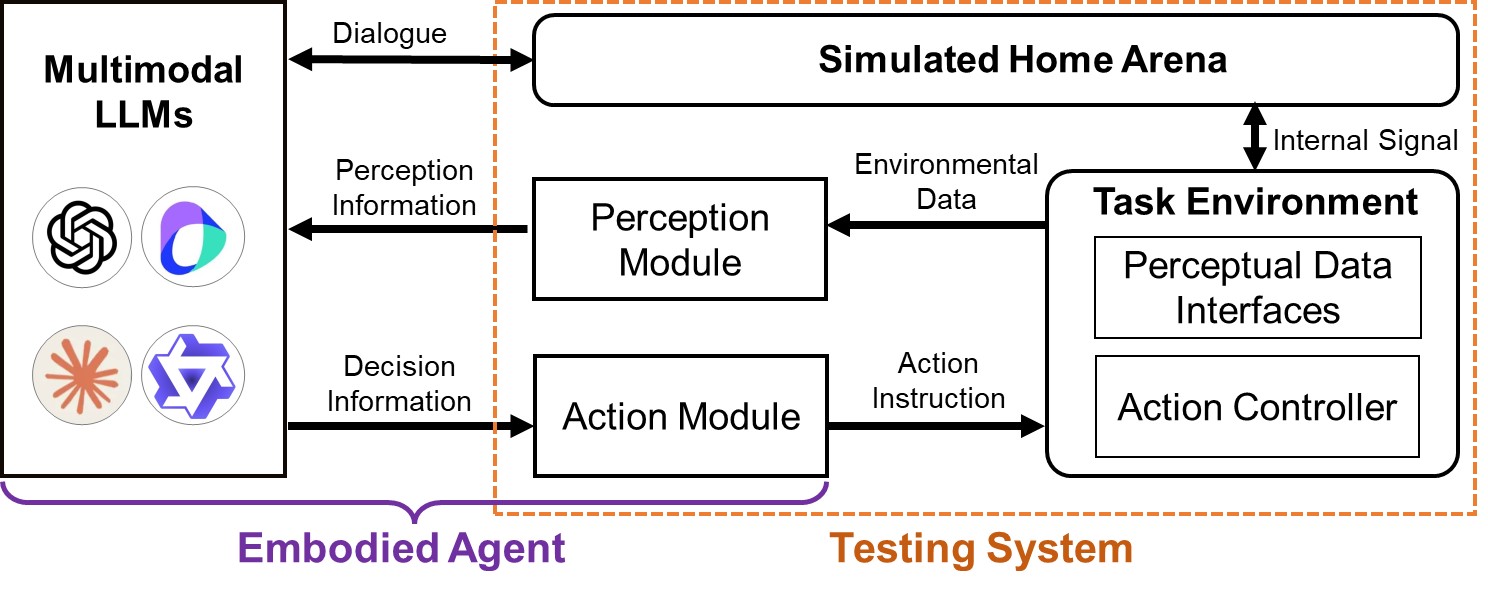}
    \caption{System framework for evaluating MLLM-based embodied agents.}
    \label{fig:agent_interface}
\end{figure}

\textbf{Framework Overview.} As shown in Fig.~\ref{fig:agent_interface}, the framework comprises five elements: (1) \textit{Perception}: acquire concatenated RGB triplets and object JSON from the testing environment; (2) \textit{Semantic Packaging}: serialize $I_t$, $\Sigma_t$, and $g$ into a structured prompt; (3) \textit{Reasoning \& Decision}: the MLLM generates ReAct-style reasoning trace $r_t$ and API call $a_t$; (4) \textit{Execution}: dispatch atomic or macro actions to the simulator; (5) \textit{Loop \& Logging}: obtain next observation and record $(o_t, r_t, a_t, o_{t+1})$.

\textbf{Action Space \& Tool Schema.} The action set $\mathcal{A}$ includes (1) \textit{Atomic actions}: e.g., \texttt{MoveForward()}, \texttt{PickUp(object)} and (2) \textit{Macro-actions}: e.g., \texttt{MoveandPickUp(object)}, either executed directly or expanded into primitives. All APIs are defined in the system prompt as JSON entries specifying \texttt{name}, \texttt{arguments}, \texttt{preconditions}, and \texttt{effects}.

\textbf{ReAct-style Inference.} At each step, the policy outputs $(r_t, a_t)$, where $r_t$ is a natural-language reasoning trace (non-executable) and $a_t$ is an executable API call. This design ensures interpretability while constraining actions to a safe, finite tool space.

\textbf{System Prompt Design.} For example, consider the task $g_0$ “You have three minutes to tidy up the dining table \texttt{BP\_DiningTable\_06\_C\_0} to prepare it for three people”, an illustrative system prompt for this task is given below, where the testing system (called ``Simulated Home Arena'') can recursively interact with the agent through language.

{\centering
\begin{tcolorbox}[
    width=\textwidth,
    colback=blue!5,
    breakable,
    colframe=blue!60!black,
    title= Prompts for Task $g_0$
]
\textbf{Simulated Home Arena:} 

You are a young girl situated in the environment. At each timestep $t$, you can execute the actions in $\mathcal{A}$ and the environmental information observable to you is represented as $o_t$. You will be asked to complete a series of tasks. you must execute the available actions step by step until all tasks are accomplished. Your output must be a JSON array of $(r_t, a_t)$. 

The current task is: “You have three minutes to tidy up the dining table \texttt{BP\_DiningTable\_06\_C\_0} to prepare it for three people.” Your initial observation is denoted as $o_0$. No item is currently held in your hand.\\

\textbf{Agent:} 
  
$r_0$: "To tidy up the table, it is first necessary to approach it and face it. In the current scene, the table is \texttt{BP\_DiningTable\_06\_C\_0}, and the agent needs to move to its side.",

$a_0$: \texttt{MovetoObject(BP\_DiningTable\_06\_C\_0)}\\

\textbf{Simulated Home Arena:} 

Your current observation is $o_1$. No item is currently held in your hand.\\

\textbf{Agent:}

$r_1$: "I see the table, and I also notice three teacups placed on it. However, their arrangement is somewhat disorganized. I can rearrange them neatly so that the table is properly set for three people."

$a_1$: \texttt{MoveandPickUp(BP\_Cup\_Mug\_C2A3C, which\_hand=0)}\\

\dots\dots \\

\textbf{Simulated Home Arena:} 

Your current observation is $o_T$. No item is currently held in your hand.\\

\textbf{Agent:}

$r_T$: “I observe that all teacups are neatly arranged, indicating that the dining table has been successfully tidied.”

$a_T$: \texttt{finish\_task()}

\end{tcolorbox}
}

\section{Experiments}

\subsection{System Setup}

\textbf{Agent Interface.}
To support the embodied evaluation for MLLMs, we developed a virtual home environment based on Unreal Engine 5 to simulate daily activities with desired features. Each MLLM agent can exchange information with the simulated environment.
Although some MLLMs can process images and texts, they are not embodied agents without perception and action modules. In order to integrate the MLLMs into the embodied task testing system, we have developed standard perception and action modules for these MLLMs. Overall, the testing system provides standard interfaces to make it possible to test MLLMs as embodied agents, as shown in Fig.~\ref{fig:agent_interface}.

\textbf{Testing Pipeline.}
Since our testing objective is to measure the ability of the MLLM as an agent in embodied interaction environments, we only encapsulated the MLLM into a measurable agent by adding perception and action modules, without retraining the model or adding other external components.
In the specific testing process, the testing system first starts the testing environment and instantiated a virtual human as an embodiment that can be controlled by the MLLMs. Then, based on the task content, the testing system issues a task instruction described in natural language, such as "Please tidy up your room". Next, the tested MLLM-based agent will receive the instruction and generate control signals (i.e., return the APIs to call to control the virtual human), driving the virtual human to execute the task in the virtual environment, and can read the latest status from the environment at any time. The task will terminate when time runs out or when the agent issues a termination command.

\begin{table*}[b]
\centering
\caption{Model Performance Comparison}
\label{tab:model_comparison}
\resizebox{\textwidth}{!}{%
\begin{tabular}{l*{10}{c}} 
\toprule
\multirow{2}{*}{Model} & 
\multicolumn{2}{c}{Object Understanding} & 
\multicolumn{3}{c}{Spatial Intelligence} & 
\multicolumn{3}{c}{Social Activity} & 
\multirow{2}{*}{Mean} \\
\cmidrule(lr){2-3} \cmidrule(lr){4-6} \cmidrule(lr){7-9}
&  Counting Objects & Selecting Gifts & Building Blocks & Jigsaw Puzzle & Understanding Buttons & Setting Tables & 
Tidying Up Rooms & Preparing Baggage & \\
\midrule

Gemini-2.5-Pro	        &48.00 	&68.06 	&10.00 	&5.05 	&3.33 	&26.68 	&22.77 	&\textbf{12.38} 	&\textbf{24.53} \\
Gemini-2.5-Flash	    &42.00 	&68.20 	&5.50 	&5.30 	&3.33 	&25.83 	&\textbf{23.22} 	&11.05 	&23.05 \\
o3	                    &\textbf{54.00}  &65.92 	&10.00 	&6.40 	&3.33 	&14.31 	&18.77 	&10.30 	&22.88 \\
GPT-5	                &36.00 	&\textbf{69.06} 	&3.75 	&6.03 	&3.33 	&\textbf{28.68} 	&16.00 	&9.50 	&21.54 \\
Claude-3.7-Sonnet	    &46.00 	&59.74 	&8.88 	&6.28 	&0.00 	&23.76 	&16.12 	&3.40 	&20.52 \\
Claude-4-Sonnet	        &44.00 	&65.44 	&8.75 	&6.03 	&0.00 	&19.81 	&14.85 	&5.18 	&20.51 \\
Doubao-1.5-vision-pro	&36.00 	&56.70 	&\textbf{12.50} 	&6.88 	&\textbf{5.00} 	&24.17 	&4.22 	&7.75 	&19.15 \\
Claude-3.5-Sonnet	    &42.00 	&64.24 	&8.00 	&4.50 	&0.00 	&12.33 	&13.47 	&6.00 	&18.82 \\
Grok 3	                &52.00 	&50.24 	&9.88 	&\textbf{8.00} 	&0.00 	&8.37 	&17.20 	&3.35 	&18.63 \\
o4-mini	                &38.00 	&66.40 	&0.00 	&6.62 	&3.33 	&13.67 	&1.52 	&4.23 	&16.72 \\
GPT-4o 	                &32.00 	&46.30 	&10.00 	&7.08 	&3.33 	&18.42 	&6.30 	&8.18 	&16.45 \\
GPT-4o-mini	            &34.00 	&54.96 	&5.75 	&7.05 	&1.67 	&19.07 	&1.50 	&3.75 	&15.97 \\
Qwen-VL-max	            &44.00 	&48.14 	&0.00 	&7.55 	&0.00 	&15.96 	&1.25 	&0.00 	&14.61 \\
Llama-4-Maverick	    &36.00 	&50.30 	&3.75 	&6.58 	&0.00 	&15.75 	&0.42 	&3.00 	&14.48 \\
Llama-4-Scout	        &28.00 	&42.56 	&6.25 	&5.72 	&0.00 	&8.94 	&1.38 	&2.00 	&11.86 \\
Qwen-VL-plus	        &6.00 	&34.16 	&0.00 	&7.65 	&0.00 	&16.67 	&1.82 	&0.38 	&8.33 \\
Llama-3.2	            &6.00 	&6.80 	&0.00 	&4.60 	&0.00 	&4.67 	&0.42 	&0.00 	&2.81 \\

\bottomrule
\end{tabular}%
}
\end{table*}

\subsection{Results}

\begin{figure}[t]
    \centering
    \includegraphics[width=1\linewidth]{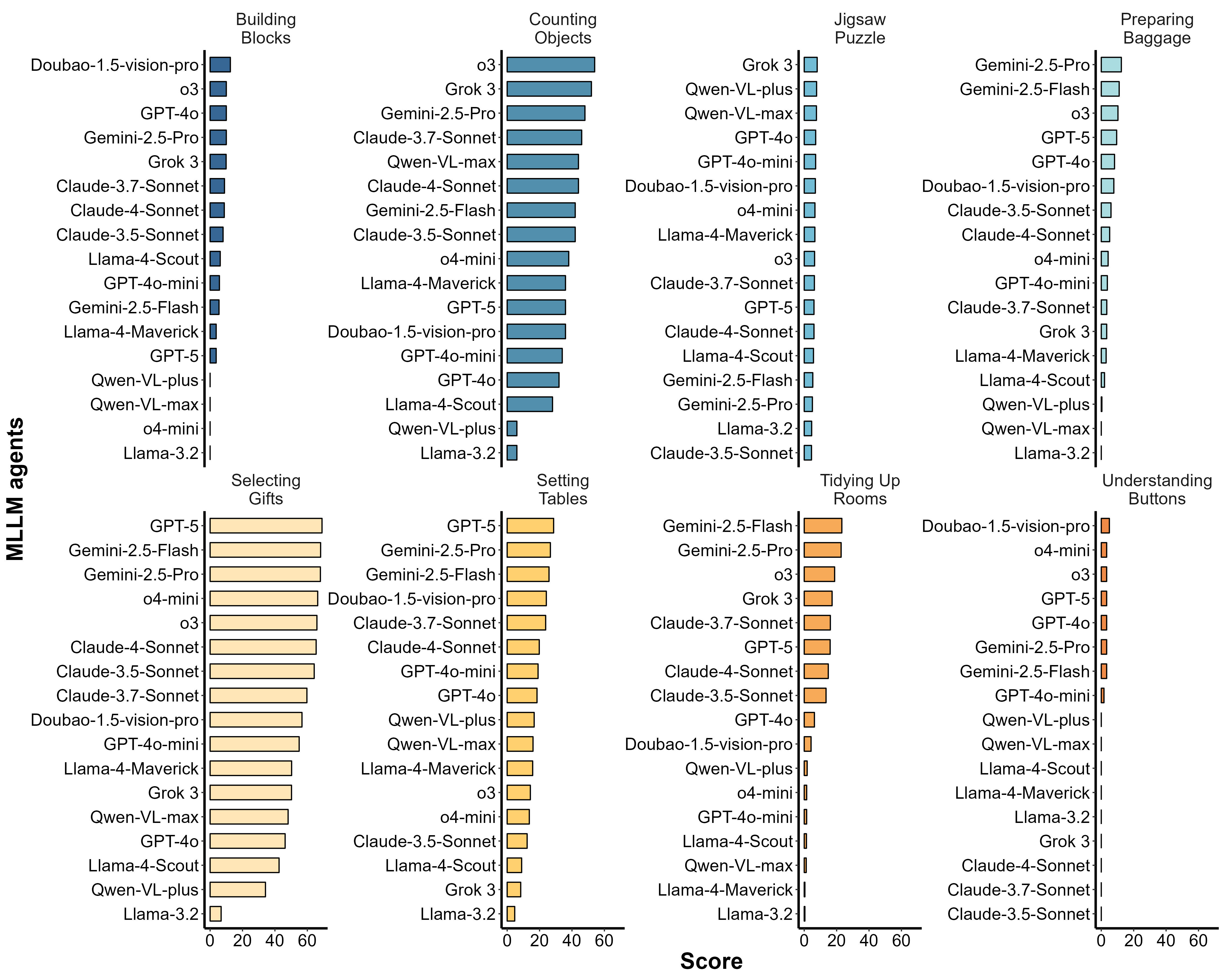}
    \caption{MLLM agents' performance on eight tasks types. }
    \vspace{-6pt}
    \label{fig:eighttasks}
\end{figure}

\textbf{Overall Results.} Table. \ref{tab:model_comparison}  and Fig. \ref{fig:eighttasks} present the evaluation results in eight embodied tasks, along with the average performance. Overall, MLLM agents demonstrated a lack of ability to perform composite tasks grounded in home environments. The highest average score was achieved by Gemini-2.5-Pro \cite{team2023gemini}, reaching only 24.53 out of 100. The proprietary models generally outperformed the open-source ones, except Qwen-VL-plus\cite{bai2023versatile}, which fell between that of the open-source models with a score of 8.33. Among the open-source models, the highest average score was 14.48 by Llama-4-Maverick \cite{huggingface2025llama4}. Nevertheless, the performance gaps between the proprietary and open-source models were not substantial. These results suggest that while current MLLMs demonstrate promise in perceptual recognition, their architectures, training regimes, and multimodal integration strategies may still be insufficient for achieving general-purpose embodied intelligence.

\begin{wrapfigure}{r}{0.49\textwidth}
    \centering
    \includegraphics[width=0.485\textwidth]{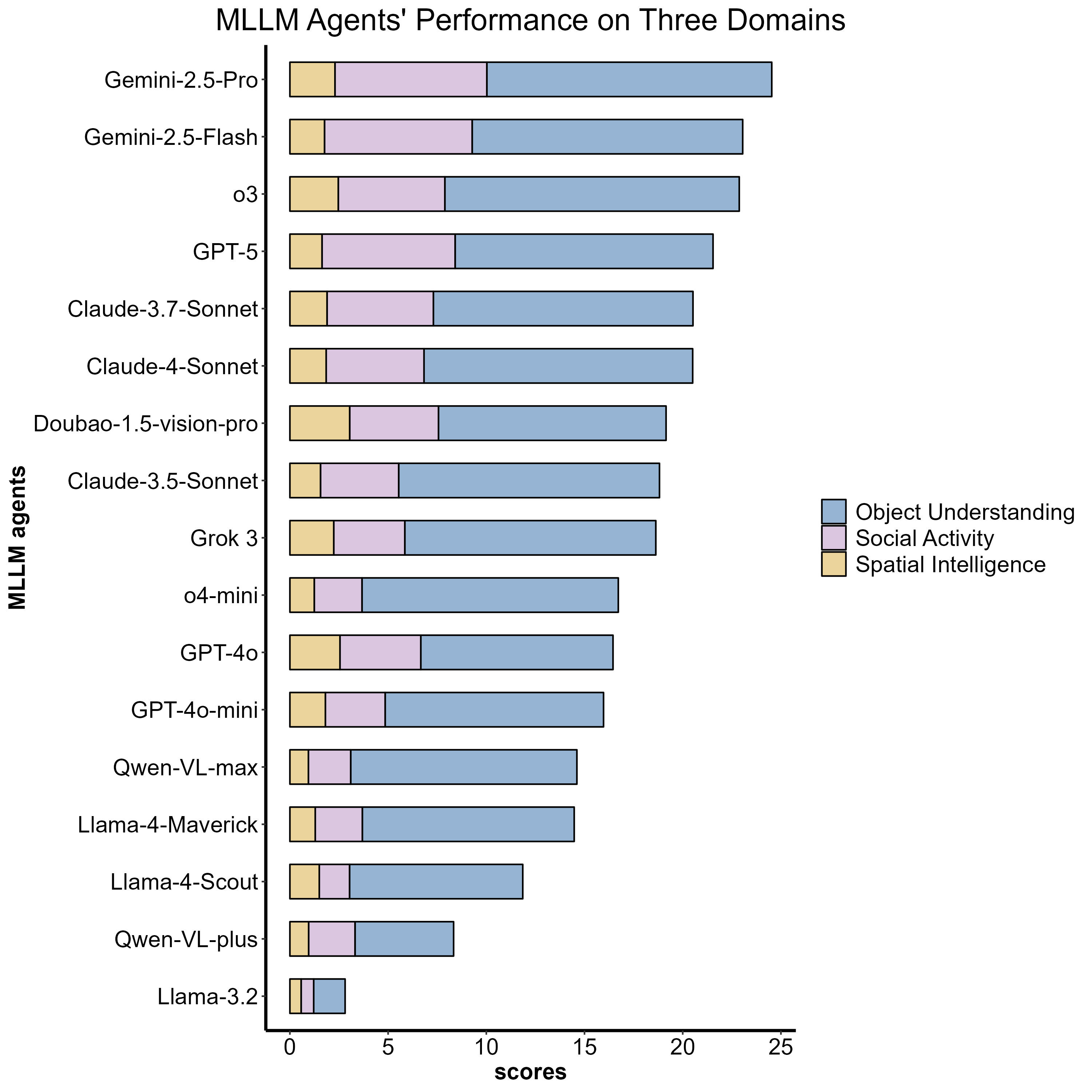}
    \caption{MLLM agents' performance on three domains. }  \vspace{-24pt}
    \label{fig:threedomains}
\end{wrapfigure}

\begin{figure*}[htbp]
    \centering
    \begin{subfigure}[b]{0.32\textwidth} 
        \centering
        \includegraphics[width=\linewidth]{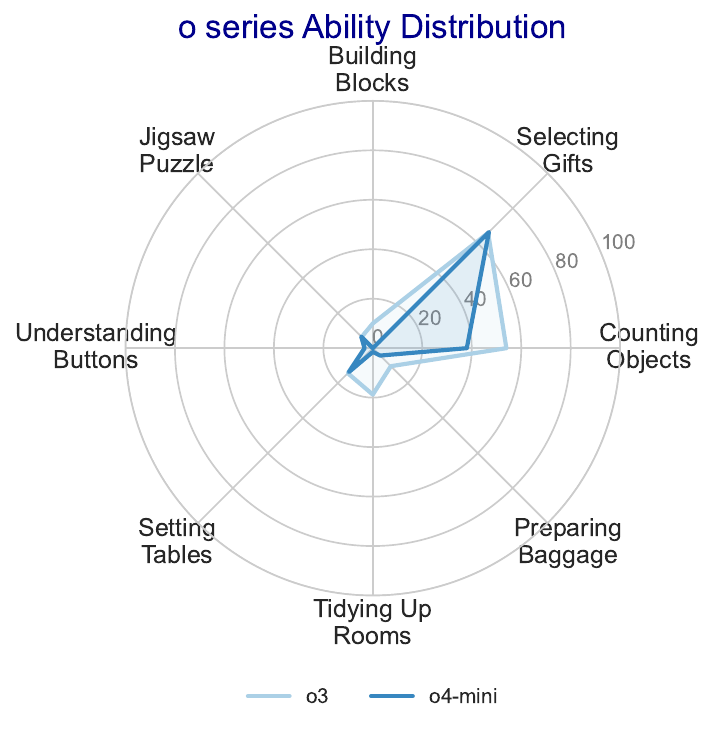} 
        \caption{o series} 
        \label{fig:radar1}
    \end{subfigure}
    \hfill 
    \begin{subfigure}[b]{0.32\textwidth} 
        \centering
        \includegraphics[width=\linewidth]{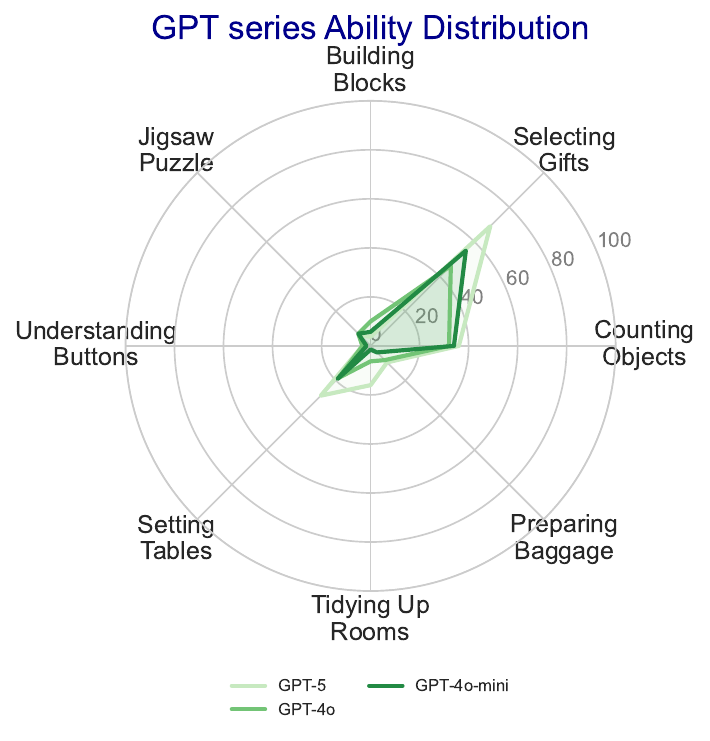} 
        \caption{GPT series} 
        \label{fig:radar2}
    \end{subfigure}
    \hfill 
    \begin{subfigure}[b]{0.32\textwidth} 
        \centering
        \includegraphics[width=\linewidth]{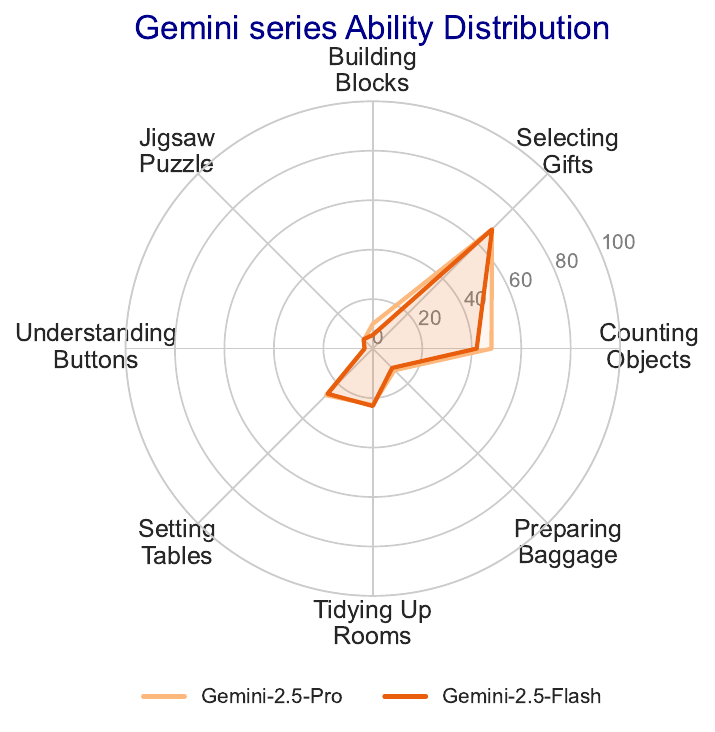} 
        \caption{Gemini series} 
        \label{fig:radar3}
    \end{subfigure}
    \hfill 
        \begin{subfigure}[b]{0.32\textwidth} 
        \centering
        \includegraphics[width=\linewidth]{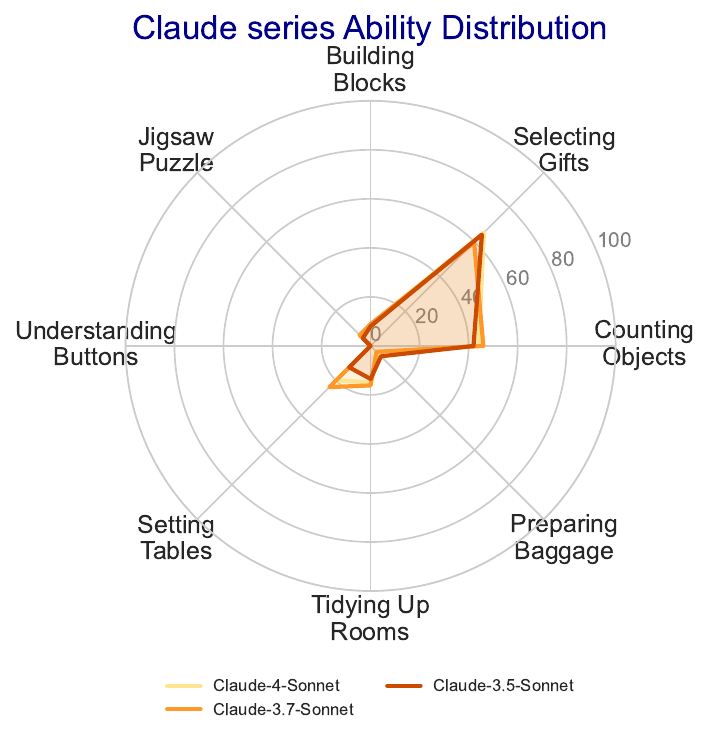} 
        \caption{Claude series} 
        \label{fig:radar4}
    \end{subfigure}
    \hfill
    \begin{subfigure}[b]{0.32\textwidth} 
        \centering
        \includegraphics[width=\linewidth]{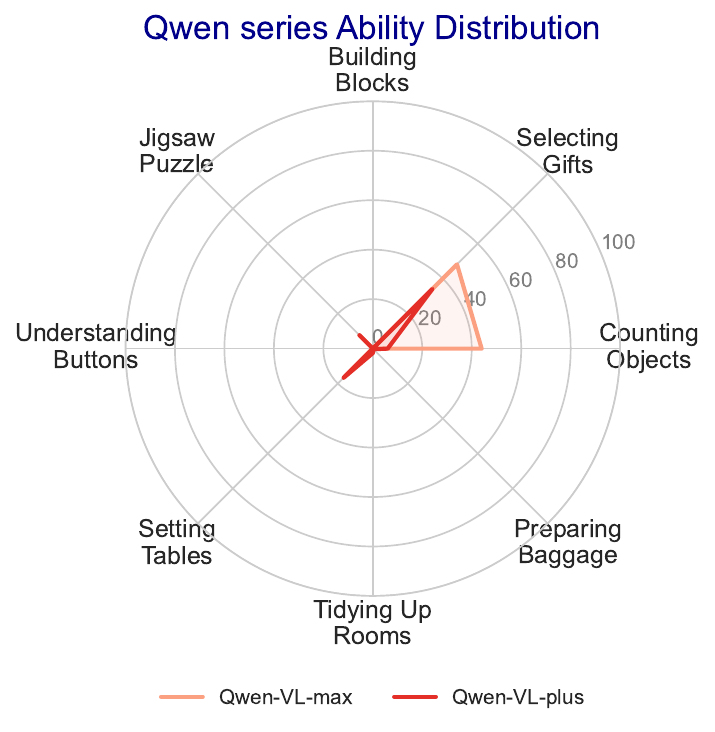} 
        \caption{Qwen series} 
        \label{fig:radar5}
    \end{subfigure}
    \hfill 
        \begin{subfigure}[b]{0.32\textwidth} 
        \centering
        \includegraphics[width=\linewidth]{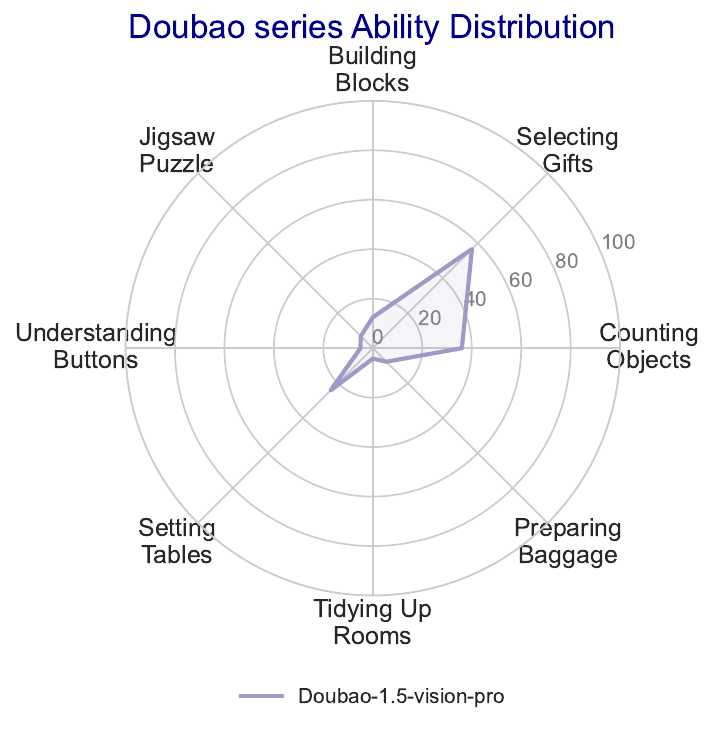} 
        \caption{Doubao 1.5} 
        \label{fig:radar6}
    \end{subfigure}
    \hfill 
    \begin{subfigure}[b]{0.32\textwidth} 
        \centering
        \includegraphics[width=\linewidth]{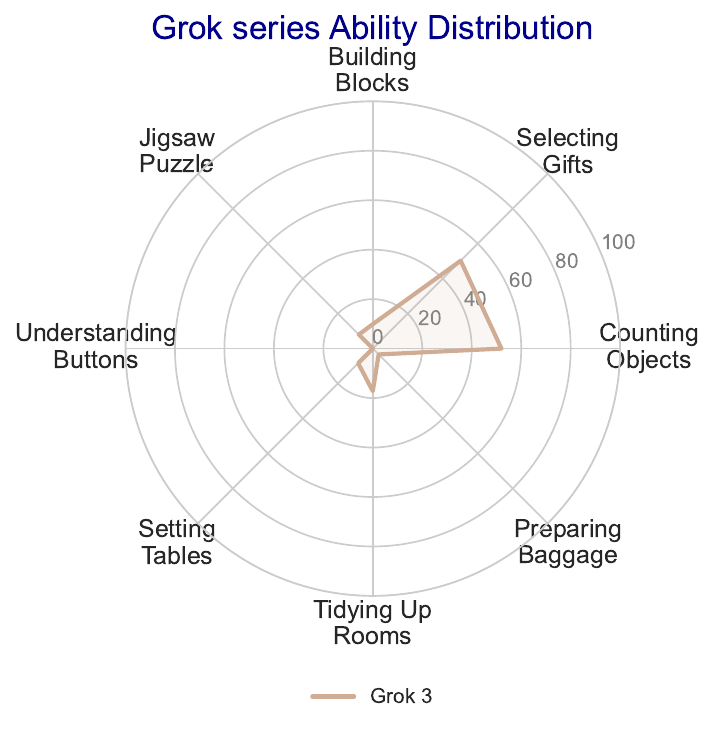} 
        \caption{Grok 3} 
        \label{fig:radar7}
    \end{subfigure}
    \hfill 
    \begin{subfigure}[b]{0.32\textwidth} 
        \centering
        \includegraphics[width=\linewidth]{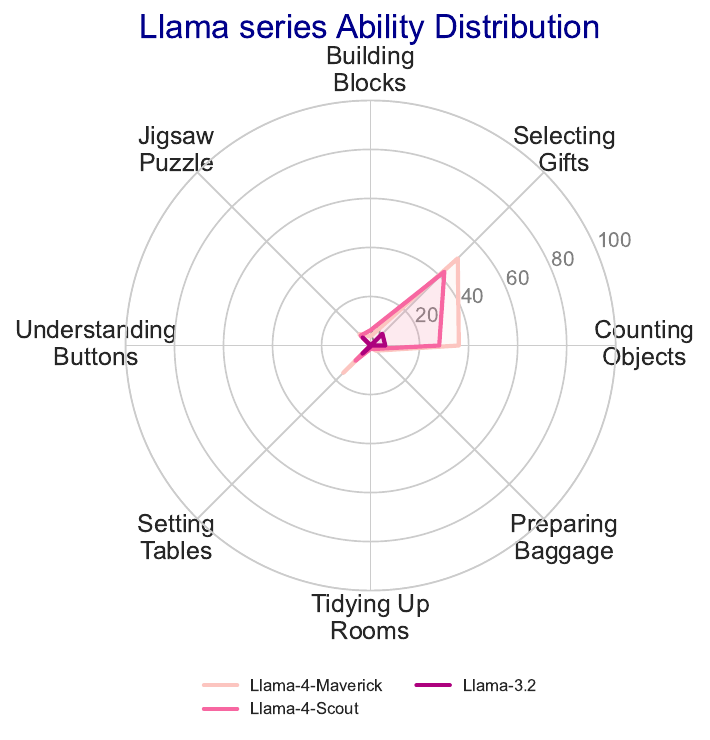} 
        \caption{Llama series} 
        \label{fig:radar8}
    \end{subfigure}
    \hspace{0.32\textwidth}
    \caption{Ability distribution of 8 series of MLLM agents across eight task categories.} 
    \label{fig:all_radars_series}
\end{figure*}

\textbf{Fine-Grained Performance Across Domains.} Breaking down the results into the three core domains, most models demonstrated relatively stronger performance in object understanding compared to spatial intelligence and social activity, as shown in Fig. \ref{fig:threedomains} and Fig. \ref{fig:all_radars_series}. These results suggest that while current MLLM agents retain some strength in perceptual recognition tasks, they continue to struggle with more integrated, real-world embodied intelligence.

\textbf{Performance varied across tasks and models.} In the object understanding domain, o3 \cite{openai_o3_o4mini_2025} achieved the highest score for counting objects (54.00), while GPT-5  \cite{openai_gpt5_2025} led in selecting gifts (69.06). In contrast, Llama-3.2 performed the worst in both subtasks and consistently showed low scores in seven of the eight tasks, except for the Jigsaw Puzzle. This underperformance likely stems from its limitations in precise object recognition, classification, and contextual understanding. 

In the spatial intelligence domain, Doubao-1.5-vision-pro \cite{volcengine2025doubao} achieved the highest score in building blocks (12.50) and understanding buttons (5.00), while Grok 3 led in jigsaw puzzle (8.00). These findings indicate that spatial reasoning and manipulation remain particularly challenging for most MLLM embodied agents.

In the domain of social activity, different models excelled in different tasks: GPT-5 in setting tables (28.68), Gemini-2.5-Flash in tidying up rooms (23.22), and Gemini-2.5-Pro in preparing baggage (12.38). However, despite leading their respective tasks, these scores remain limited. These results indicate that even the most advanced MLLMs struggle with complex and goal-directed social understanding, and no current model demonstrates robust competence in this domain.

\section{Discussion}

The current study proposed a set of composite tasks inspired by common daily activities within a dynamic and simulated home environment, covering three categories: object understanding, spatial intelligence, and social activity. A total of 17 leading proprietary and open-source MLLMs were evaluated based on composite tasks. Our results demonstrated that MLLM agents perform poorly on these tasks, suggesting broader challenges in dynamic spatial reasoning and contextually appropriate social activities. The limitations were systemic across different MLLMs, rather than confined to specific model families. 

Although recent MLLMs demonstrate progress  in perceptual recognition, they may still lack the robust multimodal integration and embodied reasoning required for general-purpose embodied intelligence. One possible explanation is that current MLLMs rely on surface-level correlations across language, perception, and motion, rather than developing a grounded conceptual understanding of objects, space, and social dynamics. Furthermore, inherent variability and randomness in task execution may amplify their lack of robustness and adaptability.  To bridge this gap, future work may focus on advancing multimodal perception, embodied reasoning, and social understanding, while also developing more precise evaluation methods to distinguish genuine general abilities from superficial task adaptation. 

This study evaluated the capabilities of MLLMs in performing common household tasks. To evaluate their native capabilities as intelligent agents, our methodology relied exclusively on prompt engineering to facilitate embodied interactions. We eschewed external agent frameworks and auxiliary modules, including dedicated memory systems, database retrieval mechanisms, and tool-calling interfaces. Although this minimalist approach may limit peak performance, it ensured a direct assessment of the MLLM's foundational capabilities, providing a critical benchmark and a valuable reference for researchers and developers considering the practical deployment of MLLMs.

The current study bears a few limitations. A key limitation of this study is that our evaluations were conducted exclusively in simulated environments, without validation through real-world experiments. This limitation arises because our tasks were drawn from common daily activities in early childhood, which were assumed to be reliably solvable by humans.  As a result, we focused exclusively on evaluating and comparing the performance of MLLM agents within a controlled setting.  Future work could extend this research by incorporating more realistic and complex simulations or by developing standardized, cost-effective real-world test suites. These efforts would help bridge the gap between experimental evaluation and practical deployment, making performance insights more applicable to real-world scenarios. In addition, we did not standardize the difficulty levels across tasks, treating each task as independent of the others. This lack of consistency may have introduced evaluation bias, as models might exhibit better performance on simpler tasks while struggling more with tasks of higher complexity. Future research should aim to align task difficulty levels to enable a more fair and meaningful comparison of model performance. Furthermore, incorporating dynamic task difficulty based on agent performance could provide deeper insight into the models' learning and adaptation capabilities.

\section{Conclusion}

The current work introduced a set of composite embodied tasks to assess the general abilities of MLLM agents in a simulated home environment. The results revealed that the current MLLMs struggled with composite tasks associated with object understanding, spatial intelligence, and social activity, underscoring the gaps between the current capabilities of the foundational models and the demands of general intelligence. Grounded in early childhood development, our task sets provide a preliminary framework for probing the general abilities of embodied agents, and may help identify critical limitations that future work should address to bring embodied MLLMs closer to real-world applications.

\section{Acknowledgments}
This work was supported by the National Science and Technology Major Project (2022ZD0114900), the State Key Laboratory of General Artificial Intelligence.

\newpage

\begin{thebibliography}{10}

\bibitem{achiam2023gpt}
J.~Achiam, S.~Adler, S.~Agarwal, L.~Ahmad, I.~Akkaya, F.~L. Aleman, D.~Almeida, J.~Altenschmidt, S.~Altman, S.~Anadkat, et~al.
\newblock Gpt-4 technical report.
\newblock {\em arXiv preprint arXiv:2303.08774}, 2023.

\bibitem{antol2015vqa}
S.~Antol, A.~Agrawal, J.~Lu, M.~Mitchell, D.~Batra, C.~L. Zitnick, and D.~Parikh.
\newblock Vqa: Visual question answering.
\newblock In {\em Proceedings of the IEEE international conference on computer vision}, pages 2425--2433, 2015.

\bibitem{bai2023versatile}
J.~Bai, S.~Bai, S.~Yang, S.~Wang, S.~Tan, P.~Wang, J.~Lin, C.~Zhou, and J.~Q.-V. Zhou.
\newblock A versatile vision-language model for understanding, localization, text reading, and beyond.
\newblock {\em arXiv preprint arXiv:2308.12966}, 6, 2023.

\bibitem{blakey2019causality}
E.~Blakey, E.~C. Tecwyn, T.~McCormack, D.~A. Lagnado, C.~Hoerl, S.~Lorimer, and M.~J. Buehner.
\newblock When causality shapes the experience of time: Evidence for temporal binding in young children.
\newblock {\em Developmental science}, 22(3):e12769, 2019.

\bibitem{brown2020language}
T.~Brown, B.~Mann, N.~Ryder, M.~Subbiah, J.~D. Kaplan, P.~Dhariwal, A.~Neelakantan, P.~Shyam, G.~Sastry, A.~Askell, et~al.
\newblock Language models are few-shot learners.
\newblock {\em Advances in neural information processing systems}, 33:1877--1901, 2020.

\bibitem{cheng2025embodiedeval}
Z.~Cheng, Y.~Tu, R.~Li, S.~Dai, J.~Hu, S.~Hu, J.~Li, Y.~Shi, T.~Yu, W.~Chen, et~al.
\newblock Embodiedeval: Evaluate multimodal llms as embodied agents.
\newblock {\em arXiv preprint arXiv:2501.11858}, 2025.

\bibitem{chollet2019measure}
F.~Chollet.
\newblock On the measure of intelligence.
\newblock {\em arXiv preprint arXiv:1911.01547}, 2019.

\bibitem{chollet2024abstraction}
F.~Chollet.
\newblock Abstraction and reasoning corpus for artificial general intelligence (arc-agi), 2024.

\bibitem{deng2009imagenet}
J.~Deng, W.~Dong, R.~Socher, L.-J. Li, K.~Li, and L.~Fei-Fei.
\newblock Imagenet: A large-scale hierarchical image database.
\newblock In {\em Proceedings of the IEEE conference on computer vision and pattern recognition}, pages 248--255. Ieee, 2009.

\bibitem{driess2023palm}
D.~Driess, F.~Xia, M.~S. Sajjadi, C.~Lynch, A.~Chowdhery, A.~Wahid, J.~Tompson, Q.~Vuong, T.~Yu, W.~Huang, et~al.
\newblock Palm-e: An embodied multimodal language model.
\newblock 2023.

\bibitem{huggingface2025llama4}
H.~Face.
\newblock Meta llama 4 collection, 2025.
\newblock Accessed: 2025-09-19.

\bibitem{gan2020threedworld}
C.~Gan, J.~Schwartz, S.~Alter, D.~Mrowca, M.~Schrimpf, J.~Traer, J.~De~Freitas, J.~Kubilius, A.~Bhandwaldar, N.~Haber, et~al.
\newblock Threedworld: A platform for interactive multi-modal physical simulation.
\newblock {\em arXiv preprint arXiv:2007.04954}, 2020.

\bibitem{gao2019vrkitchen}
X.~Gao, R.~Gong, T.~Shu, X.~Xie, S.~Wang, and S.-C. Zhu.
\newblock Vrkitchen: an interactive 3d virtual environment for task-oriented learning.
\newblock {\em arXiv preprint arXiv:1903.05757}, 2019.

\bibitem{gao2024multi}
Z.~Gao, B.~Zhang, P.~Li, X.~Ma, T.~Yuan, Y.~Fan, Y.~Wu, Y.~Jia, S.-C. Zhu, and Q.~Li.
\newblock Multi-modal agent tuning: Building a vlm-driven agent for efficient tool usage.
\newblock {\em arXiv preprint arXiv:2412.15606}, 2024.

\bibitem{he2025flagevalmm}
Z.~He, Y.~Liu, J.-s. Zheng, X.~Li, J.-G. Yao, B.~Qin, R.~Xuan, and X.~Yang.
\newblock Flagevalmm: A flexible framework for comprehensive multimodal model evaluation.
\newblock {\em arXiv preprint arXiv:2506.09081}, 2025.

\bibitem{khanna2024goatbenchbenchmarkmultimodallifelong}
M.~Khanna, R.~Ramrakhya, G.~Chhablani, S.~Yenamandra, T.~Gervet, M.~Chang, Z.~Kira, D.~S. Chaplot, D.~Batra, and R.~Mottaghi.
\newblock Goat-bench: A benchmark for multi-modal lifelong navigation, 2024.

\bibitem{kirillov2023segment}
A.~Kirillov, E.~Mintun, N.~Ravi, H.~Mao, C.~Rolland, L.~Gustafson, T.~Xiao, S.~Whitehead, A.~C. Berg, W.-Y. Lo, et~al.
\newblock Segment anything.
\newblock In {\em Proceedings of the IEEE/CVF international conference on computer vision}, pages 4015--4026, 2023.

\bibitem{kolve2017ai2}
E.~Kolve, R.~Mottaghi, W.~Han, E.~VanderBilt, L.~Weihs, A.~Herrasti, M.~Deitke, K.~Ehsani, D.~Gordon, Y.~Zhu, et~al.
\newblock Ai2-thor: An interactive 3d environment for visual ai.
\newblock {\em arXiv preprint arXiv:1712.05474}, 2017.

\bibitem{levine2012early}
S.~C. Levine, K.~R. Ratliff, J.~Huttenlocher, and J.~Cannon.
\newblock Early puzzle play: a predictor of preschoolers' spatial transformation skill.
\newblock {\em Developmental psychology}, 48(2):530, 2012.

\bibitem{li202511plus}
C.~Li, W.~Wu, H.~Zhang, Q.~Li, Z.~Gao, Y.~Xia, J.~Hern{\'a}ndez-Orallo, I.~Vuli{\'c}, and F.~Wei.
\newblock 11plus-bench: Demystifying multimodal llm spatial reasoning with cognitive-inspired analysis.
\newblock {\em arXiv preprint arXiv:2508.20068}, 2025.

\bibitem{li2023behavior}
C.~Li, R.~Zhang, J.~Wong, C.~Gokmen, S.~Srivastava, R.~Mart{\'\i}n-Mart{\'\i}n, C.~Wang, G.~Levine, M.~Lingelbach, J.~Sun, et~al.
\newblock Behavior-1k: A benchmark for embodied ai with 1,000 everyday activities and realistic simulation.
\newblock In {\em Conference on Robot Learning}, pages 80--93. PMLR, 2023.

\bibitem{li2024embodied}
M.~Li, S.~Zhao, Q.~Wang, K.~Wang, Y.~Zhou, S.~Srivastava, C.~Gokmen, T.~Lee, E.~L. Li, R.~Zhang, et~al.
\newblock Embodied agent interface: Benchmarking llms for embodied decision making.
\newblock {\em Advances in Neural Information Processing Systems}, 37:100428--100534, 2024.

\bibitem{lin2014microsoft}
T.-Y. Lin, M.~Maire, S.~Belongie, J.~Hays, P.~Perona, D.~Ramanan, P.~Doll{\'a}r, and C.~L. Zitnick.
\newblock Microsoft coco: Common objects in context.
\newblock In {\em European conference on computer vision}, pages 740--755. Springer, 2014.

\bibitem{liu2024mmbench}
Y.~Liu, H.~Duan, Y.~Zhang, B.~Li, S.~Zhang, W.~Zhao, Y.~Yuan, J.~Wang, C.~He, Z.~Liu, et~al.
\newblock Mmbench: Is your multi-modal model an all-around player?
\newblock In {\em European conference on computer vision}, pages 216--233. Springer, 2024.

\bibitem{morris2024position}
M.~R. Morris, J.~Sohl-Dickstein, N.~Fiedel, T.~Warkentin, A.~Dafoe, A.~Faust, C.~Farabet, and S.~Legg.
\newblock Position: Levels of agi for operationalizing progress on the path to agi.
\newblock In {\em Forty-first International Conference on Machine Learning}, 2024.

\bibitem{mulyana2022effect}
A.~Mulyana and N.~Nurcahyani.
\newblock The effect of the puzzle playing method on improving the cognitive development of children aged 4-6 years.
\newblock {\em KnE Life Sciences}, pages 542--548, 2022.

\bibitem{nie2025wmnav}
D.~Nie, X.~Guo, Y.~Duan, R.~Zhang, and L.~Chen.
\newblock Wmnav: Integrating vision-language models into world models for object goal navigation.
\newblock {\em arXiv preprint arXiv:2503.02247}, 2025.

\bibitem{openai_gpt5_2025}
OpenAI.
\newblock Gpt-5 system card, Aug. 2025.
\newblock Accessed: 2025-08-13.

\bibitem{openai_o3_o4mini_2025}
OpenAI.
\newblock Openai o3 and o4-mini system card, Apr. 2025.
\newblock Accessed: 2025-04-16.

\bibitem{oppy2003turing}
G.~Oppy and D.~Dowe.
\newblock The turing test.
\newblock 2003.

\bibitem{peng2024tong}
Y.~Peng, J.~Han, Z.~Zhang, L.~Fan, T.~Liu, S.~Qi, X.~Feng, Y.~Ma, Y.~Wang, and S.-C. Zhu.
\newblock The tong test: Evaluating artificial general intelligence through dynamic embodied physical and social interactions.
\newblock {\em Engineering}, 34:12--22, 2024.

\bibitem{puig2018virtualhome}
X.~Puig, K.~Ra, M.~Boben, J.~Li, T.~Wang, S.~Fidler, and A.~Torralba.
\newblock Virtualhome: Simulating household activities via programs.
\newblock In {\em Proceedings of the IEEE conference on computer vision and pattern recognition}, pages 8494--8502, 2018.

\bibitem{puig2023habitat}
X.~Puig, E.~Undersander, A.~Szot, M.~D. Cote, T.-Y. Yang, R.~Partsey, R.~Desai, A.~W. Clegg, M.~Hlavac, S.~Y. Min, et~al.
\newblock Habitat 3.0: A co-habitat for humans, avatars and robots.
\newblock {\em arXiv preprint arXiv:2310.13724}, 2023.

\bibitem{radford2019language}
A.~Radford, J.~Wu, R.~Child, D.~Luan, D.~Amodei, I.~Sutskever, et~al.
\newblock Language models are unsupervised multitask learners.
\newblock {\em OpenAI blog}, 1(8):9, 2019.

\bibitem{sarch2024vlm}
G.~Sarch, L.~Jang, M.~Tarr, W.~W. Cohen, K.~Marino, and K.~Fragkiadaki.
\newblock Vlm agents generate their own memories: Distilling experience into embodied programs of thought.
\newblock {\em Advances in Neural Information Processing Systems}, 37:75942--75985, 2024.

\bibitem{savva2019habitat}
M.~Savva, A.~Kadian, O.~Maksymets, Y.~Zhao, E.~Wijmans, B.~Jain, J.~Straub, J.~Liu, V.~Koltun, J.~Malik, et~al.
\newblock Habitat: A platform for embodied ai research.
\newblock In {\em Proceedings of the IEEE/CVF international conference on computer vision}, pages 9339--9347, 2019.

\bibitem{scharf2016developmental}
R.~J. Scharf, G.~J. Scharf, and A.~Stroustrup.
\newblock Developmental milestones.
\newblock {\em Pediatrics in review}, 37(1):25--38, 2016.

\bibitem{sheldrick2019establishing}
R.~C. Sheldrick, L.~E. Schlichting, B.~Berger, A.~Clyne, P.~Ni, E.~C. Perrin, and P.~M. Vivier.
\newblock Establishing new norms for developmental milestones.
\newblock {\em Pediatrics}, 144(6):e20190374, 2019.

\bibitem{shridhar2020alfred}
M.~Shridhar, J.~Thomason, D.~Gordon, Y.~Bisk, W.~Han, R.~Mottaghi, L.~Zettlemoyer, and D.~Fox.
\newblock Alfred: A benchmark for interpreting grounded instructions for everyday tasks.
\newblock In {\em Proceedings of the IEEE/CVF conference on computer vision and pattern recognition}, pages 10740--10749, 2020.

\bibitem{smidts2018object}
D.~P. Smidts, R.~Jacobs, and V.~Anderson.
\newblock The object classification task for children (octc): A measure of concept generation and mental flexibility in early childhood.
\newblock In {\em Using Developmental, Cognitive, and Neuroscience Approaches to Understand Executive Control in Young Children}, pages 385--401. Psychology Press, 2018.

\bibitem{spelke2007core}
E.~S. Spelke and K.~D. Kinzler.
\newblock Core knowledge.
\newblock {\em Developmental science}, 10(1):89--96, 2007.

\bibitem{szot2021habitat}
A.~Szot, A.~Clegg, E.~Undersander, E.~Wijmans, Y.~Zhao, J.~Turner, N.~Maestre, M.~Mukadam, D.~S. Chaplot, O.~Maksymets, et~al.
\newblock Habitat 2.0: Training home assistants to rearrange their habitat.
\newblock {\em Advances in neural information processing systems}, 34:251--266, 2021.

\bibitem{szot2025multimodal}
A.~Szot, B.~Mazoure, O.~Attia, A.~Timofeev, H.~Agrawal, D.~Hjelm, Z.~Gan, Z.~Kira, and A.~Toshev.
\newblock From multimodal llms to generalist embodied agents: Methods and lessons.
\newblock In {\em Proceedings of the Computer Vision and Pattern Recognition Conference}, pages 10644--10655, 2025.

\bibitem{szot2023large}
A.~Szot, M.~Schwarzer, H.~Agrawal, B.~Mazoure, W.~Talbott, K.~Metcalf, N.~Mackraz, D.~Hjelm, and A.~Toshev.
\newblock Large language models as generalizable policies for embodied tasks.
\newblock {\em arXiv preprint arXiv:2310.17722}, 2023.

\bibitem{taanila2005infant}
A.~Taanila, G.~K. Murray, J.~Jokelainen, M.~Isohanni, and P.~Rantakallio.
\newblock Infant developmental milestones: a 31-year follow-up.
\newblock {\em Developmental medicine and child neurology}, 47(9):581--586, 2005.

\bibitem{team2023gemini}
G.~Team, R.~Anil, S.~Borgeaud, J.-B. Alayrac, J.~Yu, R.~Soricut, J.~Schalkwyk, A.~M. Dai, A.~Hauth, K.~Millican, et~al.
\newblock Gemini: a family of highly capable multimodal models.
\newblock {\em arXiv preprint arXiv:2312.11805}, 2023.

\bibitem{volcengine2025doubao}
{VolcEngine}.
\newblock Doubao large language model.
\newblock \url{https://www.volcengine.com/product/doubao}, 2025.
\newblock Accessed: 2025-09-16.

\bibitem{wang2024grutopia}
H.~Wang, J.~Chen, W.~Huang, Q.~Ben, T.~Wang, B.~Mi, T.~Huang, S.~Zhao, Y.~Chen, S.~Yang, et~al.
\newblock Grutopia: Dream general robots in a city at scale.
\newblock {\em arXiv preprint arXiv:2407.10943}, 2024.

\bibitem{xie2019vrgym}
X.~Xie, H.~Liu, Z.~Zhang, Y.~Qiu, F.~Gao, S.~Qi, Y.~Zhu, and S.-C. Zhu.
\newblock Vrgym: A virtual testbed for physical and interactive ai.
\newblock In {\em Proceedings of the ACM Turing celebration conference-China}, pages 1--6, 2019.

\bibitem{yang2025embodiedbench}
R.~Yang, H.~Chen, J.~Zhang, M.~Zhao, C.~Qian, K.~Wang, Q.~Wang, T.~V. Koripella, M.~Movahedi, M.~Li, et~al.
\newblock Embodiedbench: Comprehensive benchmarking multi-modal large language models for vision-driven embodied agents.
\newblock {\em arXiv preprint arXiv:2502.09560}, 2025.

\bibitem{yang2025instructvla}
S.~Yang, H.~Li, Y.~Chen, B.~Wang, Y.~Tian, T.~Wang, H.~Wang, F.~Zhao, Y.~Liao, and J.~Pang.
\newblock Instructvla: Vision-language-action instruction tuning from understanding to manipulation.
\newblock {\em arXiv preprint arXiv:2507.17520}, 2025.

\bibitem{yin2025spatial}
B.~Yin, Q.~Wang, P.~Zhang, J.~Zhang, K.~Wang, Z.~Wang, J.~Zhang, K.~Chandrasegaran, H.~Liu, R.~Krishna, et~al.
\newblock Spatial mental modeling from limited views.
\newblock {\em arXiv preprint arXiv:2506.21458}, 2025.

\bibitem{zhang2024navid}
J.~Zhang, K.~Wang, R.~Xu, G.~Zhou, Y.~Hong, X.~Fang, Q.~Wu, Z.~Zhang, and H.~Wang.
\newblock Navid: Video-based vlm plans the next step for vision-and-language navigation.
\newblock {\em arXiv preprint arXiv:2402.15852}, 2024.

\bibitem{zhang2019symmetrical}
Z.~Zhang, C.~Wang, D.~Weng, Y.~Liu, and Y.~Wang.
\newblock Symmetrical reality: Toward a unified framework for physical and virtual reality.
\newblock In {\em Proceedings of the IEEE Conference on Virtual Reality and 3D User Interfaces (VR)}, pages 1275--1276. IEEE, 2019.

\bibitem{zhang2024emergence}
Z.~Zhang, Z.~Zhang, Z.~Jiao, Y.~Su, H.~Liu, W.~Wang, and S.-C. Zhu.
\newblock On the emergence of symmetrical reality.
\newblock In {\em Proceedings of the IEEE Conference Virtual Reality and 3D User Interfaces (VR)}, pages 639--649. IEEE, 2024.

\bibitem{zheng2022vlmbench}
K.~Zheng, X.~Chen, O.~C. Jenkins, and X.~Wang.
\newblock Vlmbench: A compositional benchmark for vision-and-language manipulation.
\newblock {\em Advances in Neural Information Processing Systems}, 35:665--678, 2022.

\bibitem{zhou2025virtual}
Q.~Zhou, H.~Zhang, X.~Lin, Z.~Zhang, Y.~Chen, W.~Liu, Z.~Zhang, S.~Chen, L.~Fang, Q.~Lyu, et~al.
\newblock Virtual community: An open world for humans, robots, and society.
\newblock {\em arXiv preprint arXiv:2508.14893}, 2025.

\end{thebibliography}

\newpage
\section*{Appendix}
\appendix

\section{Details of Tasks} \label{AP:A}

\textbf{Task 1: Counting Objects}
\begin{itemize}
    \item \textbf{Task Description}: The agent was located in a room, and it was required to walk around the room and count the number of objects with specific attributes.
    \item \textbf{Input/Output}: Inputs included a task instruction and first-person view images; outputs were text-based responses.
    \item \textbf{Metric}: Response accuracy. No response equals failure.
    \item \textbf{Contents}: 
        1) counting objects with target category;
        2) counting color categories within all objects;
        3) counting objects with target colors;
        4) counting humans with target actions;
        5) counting humans with target clothes.
\end{itemize}

\textbf{Task 2: Building Blocks}
\begin{itemize}
    \item \textbf{Task Description}: An agent is positioned in front of a table with a stack of small cubes. Its task is to assemble the blocks into a shape that matches a goal state defined by either a language instruction or a 2D image cue.
    \item \textbf{Input/Output}: Inputs included a task instruction, first-person view images, object information, and an action list; output were action sequences.
    \item \textbf{Metric}: The similarity between the final block state and the target stage, which is evaluated by an automatic scoring system or human experts.
    \item \textbf{Contents}: 
        1) building flat blocks with language descriptions;
        2) building flat blocks with image guidance;
        3) building 3D blocks with language descriptions;
        4) building 3D blocks with three-view pictures.
\end{itemize}

\textbf{Task 3: Jigsaw Puzzle}
\begin{itemize}
    \item \textbf{Task Description}: Given the target image, the agent was required to replicate the target image by manipulating the square blocks with different image patterns.
    \item \textbf{Input/Output}:  Inputs included a task instruction, first-person view images, object information, and an action list; output was action sequences.
    \item \textbf{Metric}: Each block placed in the correct position contributes to the partial credit, as defined by the task environment. 
    \item \textbf{Contents}: 
        1) natural image puzzle;
        2) multiple-object puzzle;
        3) single object puzzle;
        4) geometric puzzle.
\end{itemize}

\textbf{Task 4: Understanding Buttons}
\begin{itemize}
    \item \textbf{Task Description}: The agent was located in a room with many buttons that can be manipulated, and the agent was required to discover the function of each button through interaction. 
    \item \textbf{Input/Output}: Inputs included a task instruction, first-person view images, object information, and an action list; outputs were text-based responses. 
    \item \textbf{Metric}: Response accuracy. No response equals failure.
    \item \textbf{Contents}: 
        1) understand the buttons of the doors;
        2) understand the buttons of the fans;
        3) understand the buttons of the lights.
\end{itemize}

\textbf{Task 5: Setting Tables}
\begin{itemize}
    \item \textbf{Task Description}: Given a cluttered desktop layout, the agent needs to organize it into a reasonable target state as required.
    \item \textbf{Input/Output}: Inputs included a task instruction, first-person view images, object information, and an action list; outputs were action sequences.
    \item \textbf{Metric}: The rationality and neatness of desktop object placement are evaluated by an automatic scoring system or human experts.
    \item \textbf{Contents}: 
        1) setting dining tables;
        2) setting desks;
        3) setting tea tables.
\end{itemize}

\textbf{Task 6: Tidying Up Rooms}
\begin{itemize}
    \item \textbf{Task Description}: Given a cluttered room state, the intelligent agent needed to organize items according to the instructions.
    \item \textbf{Input/Output}:  Inputs included a task instruction, first-person view images, object information, and an action list; outputs were action sequences.
    \item \textbf{Metric}: The proportion of correctly stored items to all items that should be processed.
    \item \textbf{Contents}: 
        1) tidying up bedrooms;
        2) tidying up kitchens;
        3) tidying up living rooms;
        4) tidying up study rooms;
        5) tidying up rooms without instructions;
        6) free exploration without instructions.
\end{itemize}

\textbf{Task 7: Preparing Baggage}
\begin{itemize}
    \item \textbf{Task Description}: The agent was required to find reasonable objects and pack them into the suitcase based on the provided contextual information.
    \item \textbf{Input/Output}: Inputs included a task instruction, first-person view images, object information, and an action list; the outputs were action sequences.
    \item \textbf{Metric}: Based on the task configuration, each item is assigned a different point score, and the final total score will be normalized to within 100 according to a pre-defined formula.
    \item \textbf{Contents}: 
        1) packing for a family trip;
        2) packing for visiting a friend's house;
        3) packing for a summer camp;
        4) packing for a spring picnic.
\end{itemize}

\textbf{Task 8: Selecting Gifts}
\begin{itemize}
    \item \textbf{Task Description}: The agent is required to select gifts for given scenarios from a gift set. 
    \item \textbf{Input/Output}: Inputs included a task instruction and first-person view images; outputs were text-based responses.
    \item \textbf{Metric}: Response accuracy. No response equals failure.
    \item \textbf{Contents}: 
        1) selecting a gift for Mother's Day;
        2) selecting a gift for Dad's birthday;
        3) selecting a gift for a friend's party;
        4) selecting a gift for visiting a sick friend;
        5) selecting New Year's gifts for friends.
\end{itemize}

\end{document}